# HIGH ACCURACY LOCATION INFORMATION EXTRACTION FROM SOCIAL NETWORK TEXTS USING NATURAL LANGUAGE PROCESSING

Lossan Bonde[1] and Severin Dembele[2]

[1]Department of Applied Sciences, Adventist University of Africa, Nairobi, Kenya
[2]Laboratoire LAMDI, Universite Nazi Boni, Bobo-Dioulasso, Burkina Faso

## ABSTRACT

*Terrorism has become a worldwide plague with severe consequences for the development of nations. Besides killing innocent people daily and preventing educational activities from taking place, terrorism is also hindering economic growth. Machine Learning (ML) and Natural Language Processing (NLP) can contribute to fighting terrorism by predicting in real-time future terrorist attacks if accurate data is available. This paper is part of a research project that uses text from social networks to extract necessary information to build an adequate dataset for terrorist attack prediction. We collected a set of 3000 social network texts about terrorism in Burkina Faso and used a subset to experiment with existing NLP solutions. The experiment reveals that existing solutions have poor accuracy for location recognition, which our solution resolves. We will extend the solution to extract dates and action information to achieve the project's goal.*

## 1. INTRODUCTION

Over the past decades, the world has faced terrorist threats that have shaken the foundations of national security and global stability. Among these attacks, the deadly September 11, 2001, attack on the Twin Towers of the World Trade Center in New York remains ingrained in the memory of everyone due to the scale of the tragedy and its global impact. In response to this situation, governments worldwide have taken measures to strengthen security and fight terrorism, mobilising all available resources, including advanced technology. Scientific researchers have played a vital role in this context, especially in computer science. They realised that machine learning could help detect terrorist attacks by analysing data transmitted over the internet [1]. It is undeniable that terrorists are increasingly using social networks, forums, and instant messaging to communicate, plan attacks, and recruit new members [2]. This internet use is not reserved for terrorists, as the general public also uses the web to exchange information about terrorist activities [3]. Analysis of this data could help detect suspicious activities and anticipate potential attacks.

However, analysing these textual data presents a major challenge due to their heterogeneity and lack of integration. Raw textual data cannot be directly used to train machine learning models; many transformations are required to identify and extract useful information and then put this information in formats and structures fit for machine learning. This research project proposes a system that extracts relevant information from the internet and social network texts and organises it into structured data for machine learning. This goal cannot be achieved in a single paper; we have divided the project into three phases. The first phase, the object of this paper, is to specifically address the question of extracting location information with high accuracy. Subsequent phases will address the extraction of other types of information, with the last phase dealing with the automatic collection of texts from the internet and social network sources. In the project's current phase, we have developed a highly accurate (98% of accuracy) location recognition system, outperforming all the selected existing solutions with which it has been compared.

This work contributes towards constructing a real-time system for detecting terrorist attacks, using supervised machine learning. The final product of the whole project will be able to analyse the necessary data from various sources, including a mobile or web platform accessible to the public, as well as the internet in general and social networks in particular. The collected information will then be used to train machine learning models to detect suspicious activities and anticipate potential attacks.

Though the proposed solution is designed in the context of Burkina Faso, the system can be easily adapted for any country by changing the location names database and the language if need be.

The remaining part of the paper is organised into four sections. Section 2 explores the related works that enable us to identify existing gaps and provide foundations upon which we have built. Section 3 introduces the methodology of the research and gives details on the work that was completed in this work. Then in Section 4, the results of this work are presented and compared to others. Finally, the last section concludes the paper and provides directions for future related research works.

## 2. RELATED WORKS

Social networks have become a source of important and huge amounts of information that, if adequately mined, can be useful to valuable applications that can help monitor and control various events, processes, and natural phenomena. The information from social networks is made of text which is unstructured by nature, and applications often will need to extract from the text structured data. Virmani et al. in [4, pp. 626, 627] have identified six challenges NLP applications face in extracting information from social network text. Out of the six challenges, one is of great interest to this study: information about entities. For the specific work of this paper, Named Entity Recognition (NER) is the focal technology considered.

Since the release of ChatGPT in November 2022, it has been used in several domains, including NER applications. This research considers it worth exploring the ChatGPT capabilities in relation to the problem at hand. Consequently, we organised the literature review of this paper into two sections. The first explores NER approaches not based on ChatGPT, and the second section revolves around solutions based on ChatGPT. Each section summarises how named entity recognition has been addressed and identifies possible gaps.

### 2.1. Non-ChatGPT Approaches to Named Entity Recognition

Information extraction is one of the successful applications of Natural Language Processing (NLP). According to Khurana et al., "extracting entities such as names, places, events, dates, times, and prices is a powerful way of summarising the information relevant to a user's needs" [4]. Named Entity Recognition is a well-known approach to performing information extraction of that nature. In [5], Pinto et al. made a performance comparison between various NER solutions and established that, in general, NLP solutions tend to lose performance when applied on social network texts. Their study concluded that *OpenNLP* was the best tool for formal texts like newspaper and web pages, but TwitterNLP was identified as the best solution for social media text.

[6] introduced an interesting study which built a "Thesaurus-based Named Entity Recognition System for detecting spatio-temporal crime events in Spanish language from Twitter" system, which is specific to the Spanish language. In exploring the various studies conducted on NER applications, we observe that the solutions are either language-specific ([7], [8], [9]) or problem-specific ([10], [3], [11]).

### 2.2. ChatGPT for Named Entity Recognition

Since its release in November 2022, ChatGPT has been applied in many fields, including NER. A Google Scholar search with the key phrase "ChatGPT for Named Entity Recognition" on June 09, 2023, returns a list of 1290 entries. It is, therefore, apparent that ChatGPT has already been explored for NER systems.

We have also observed that ChatGPT is used for NER implementations in context-specific applications. Below are some of these applications:

- NER in clinical studies [12], [13]
- NER in historical documents [14]
- NER in financial text analysis [15]
- NER in the military sector [16]
- NER in the legal sector [17]
- And NER in many other areas of applications [18]

Following the trend observed in the literature review and the result of the experiment conducted, we concluded that we needed to build a specific solution for the problem of extracting location information in the context of social network texts on terrorism for the specific case of Burkina Faso.

## 3. METHODOLOGY

This research follows a three-step process to produce the desired outcome: data collection, literature review and experiment and, finally, the design and implementation of a new solution.

### 3.1. Data Collection

As stated above, the research project's final aim is to extract relevant information from social network text and structure it into an adequate format for machine learning algorithms to learn and predict terrorist attacks. In the project's first phase, the focus is on extracting location information from the social network texts. Subsequent phases will address the extraction of other required information. Figure 1 portrays an example of a social network text such as it is acquired. The text is in French, the language used in the social network texts involved in the study. A set of 3000 social network texts of this nature have been collected.

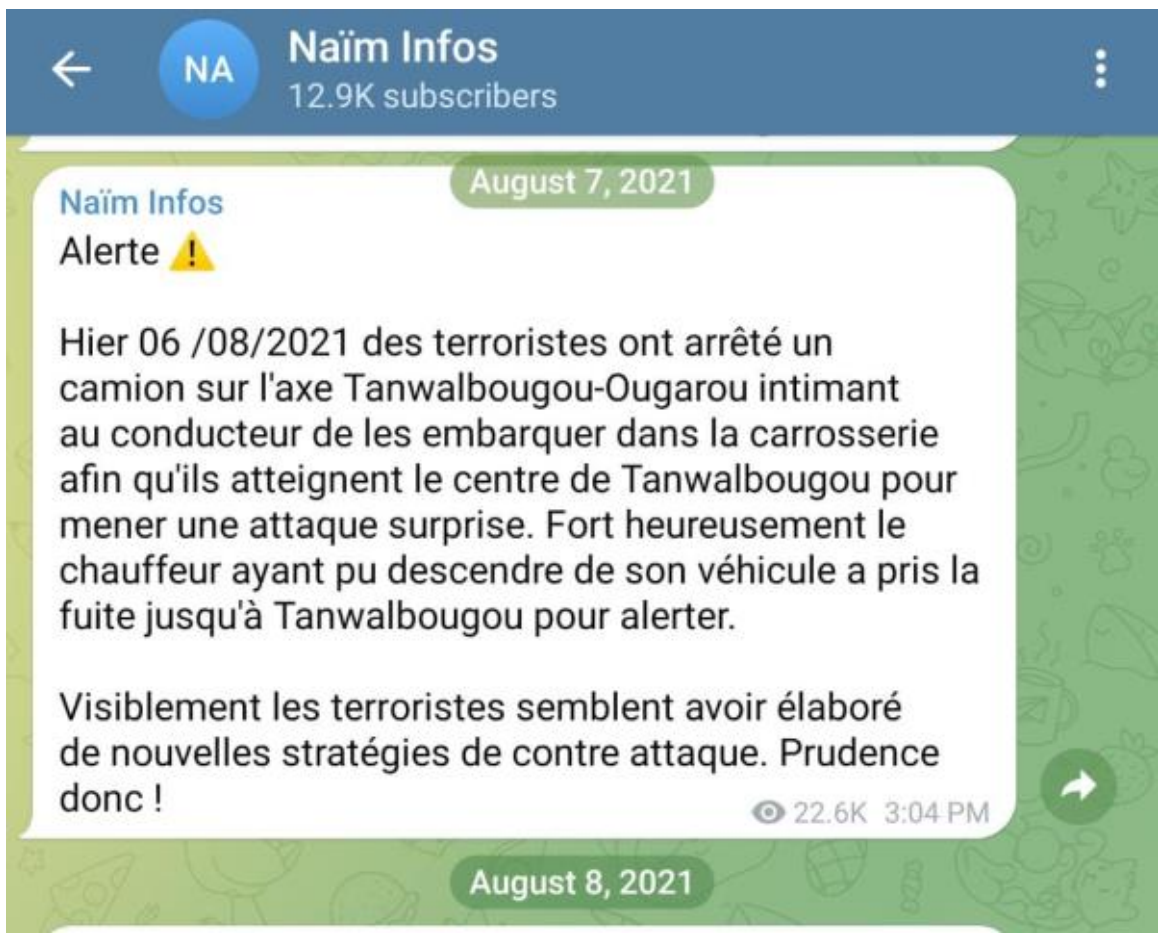

Figure 1. Example of social network text

The text in Figure 1 is a message that describes an incident that occurred on 6 August 2021, where some terrorists stopped a truck going from Tanwalbougou to Ougarou and forced the driver to take them on board. Fortunately, the driver was able to escape and raise the alert. The processing of such text should produce a list of all the locations (region, province, county, or city/village) in the text.

## 3.2. Literature Review and Experiment

With a clear knowledge of the data involved and the desired output, we did a thorough literature review to identify existing information extraction techniques. Various NER solutions were explored and, among them, ChatGPT [19], Stanford CoreNLP [20], and Spacy [21] were selected for further consideration. We conducted an experiment to assess the efficiency of these solutions. We randomly selected 20 texts out of the set of 3000 and tested each solution to determine how accurately each of them could identify location information in the texts. The details of the results are presented in Section 4. In general, the detection rates were low with the best score being 54 %. Hence, the experiment revealed that none of these existing solutions can be directly used to resolve our challenge. It was therefore necessary to design and implement a better solution, which constitutes the last step of this process.

## 3.3. Design and Implementation of a New Solution

A more accurate solution is required since the existing NLP solutions offer a poor rate of location name recognition. The approach has been to use the best of existing solutions and make some extensions that improve the recognition rate in the specific context of social network texts. To that respect, we selected Stanford CoreNLP to serve as the base NER solution. The pipeline of the proposed solution is shown in Figure 2, where two extensions have been added to the normal Stanford CoreNLP NER pipeline to consider some specific issues related to social network texts.

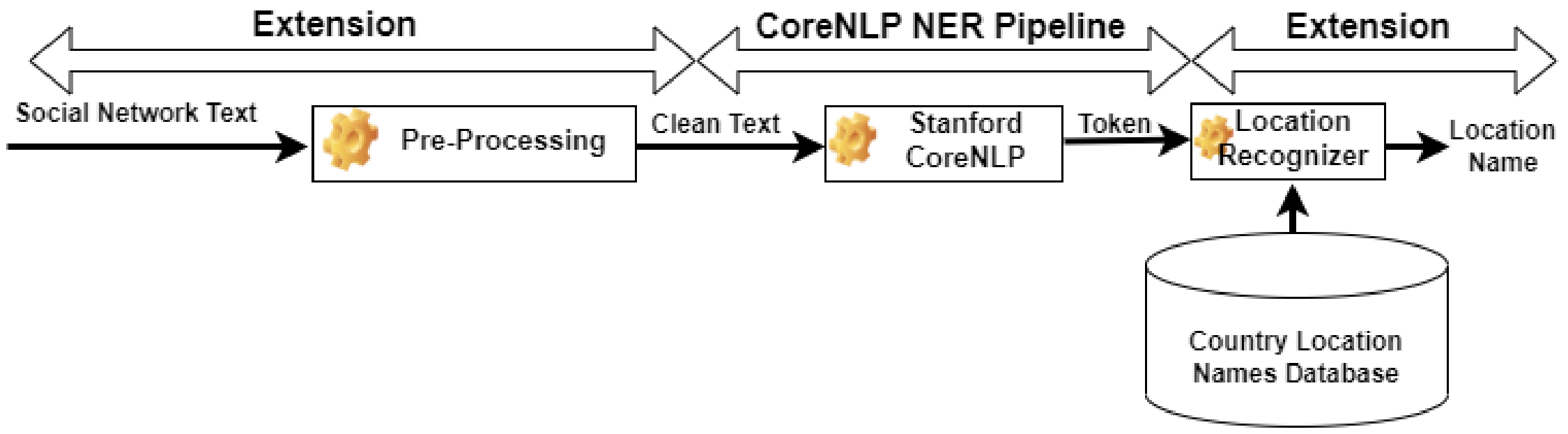

Figure 2. NER Pipeline of the Proposed Solution

### 3.3.1. CoreNLP NER Pipeline

The proposed solution uses the CoreNLP NER pipeline to split the text into tokens which are then processed by the "Location Recognizer" to identify with high accuracy location named entities. The normal CoreNLP pipeline presented in Figure 3 has been simplified because our approach only needs tokens; we stopped the pipeline just after the tokenisation step. This pipeline reduction speeds up the process.

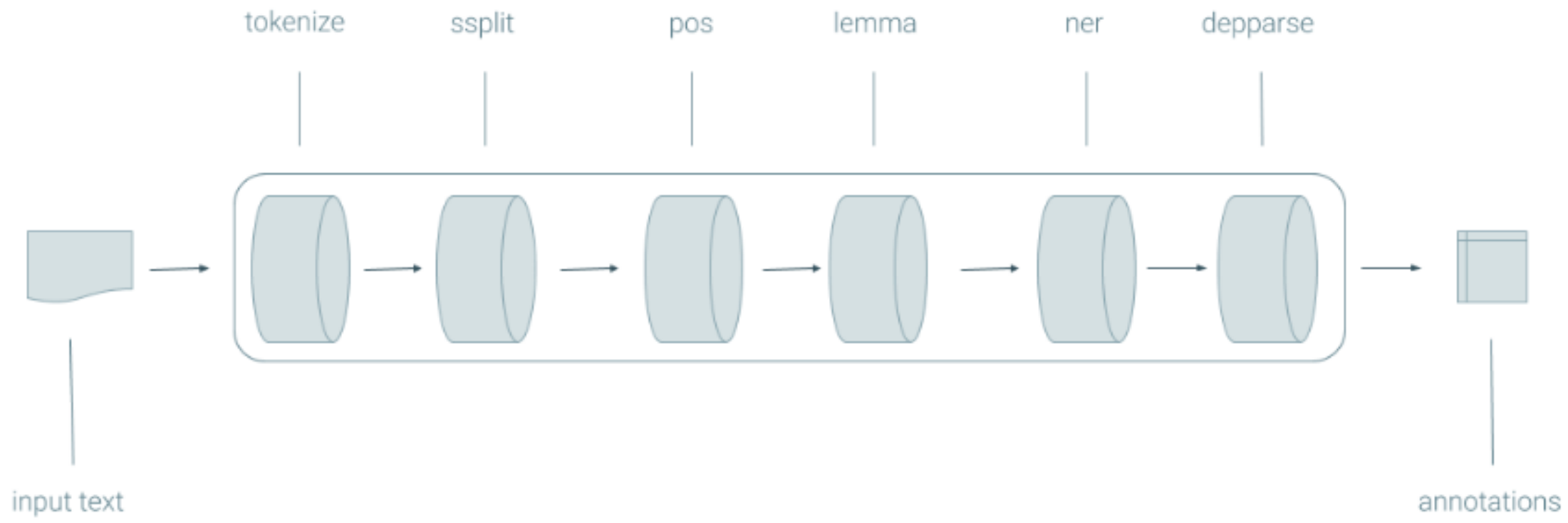

Figure 3. Stanford CoreNLP Pipeline (source:[22])

### 3.3.2. Extensions

Two extensions have been added to the normal CoreNLP pipeline: the pre-processing and the location recognizer, as shown in Figure 2 and described in the subsections below.

#### 3.3.2.1. Pre-processing

In the pre-processing extension, two types of transformations are done on the initial text: removing special symbols and normalising multi-word location names.

- **Removal of special symbols**: Social network texts, especially tweets, contain specific symbols and characters such as the hashtag (#) and the at symbol (@). The presence of those symbols can hinder the recognition of a location. The pre-processing extension removes those special symbols from the initial text before the CoreNLP process starts.
- **Normalisation of multi-word location names**: Some location names like "Bobo Dioulasso" and "Boucle du Mouhoun" are composed of multiple words, each of which will be recognised separately by the CoreNLP pipeline. During the pre-processing phase, we identify such names, and their corresponding words are combined using hyphens to make them one token in the transformed text. For example, the name "Boucle du Mouhoun" will change to "Boucle-du-Mouhoun."

After the two types of transformations, the output text called "clean text" (in the pipeline presented in Figure 2) is ready for the NER operations.

#### 3.3.2.2. Location Recognizer

The location recognizer is the heart of the proposed solution; it takes each token from the CoreNLP pipeline, looks up the database of location names, and determines if the token (the token's text) matches a known location name. If so, that name is returned as output; otherwise, the token does not correspond to a location name.

The matching system is based on the Generalized Levenshtein Distance (GLD), which measures the difference between two strings. As raised by [23], social network texts are often written with spelling errors and non-standard abbreviations. Any good NLP tool dealing with this type of text must use a string similarity algorithm to handle the misspelling occurrences. Among the multiple existing algorithms to solve this problem, Yujian and Bo [24] recommend the GLD as a better solution, which we also adopted in this research. Using the GLD algorithm and the database of the location names, the matching algorithm is depicted in Figure 4.

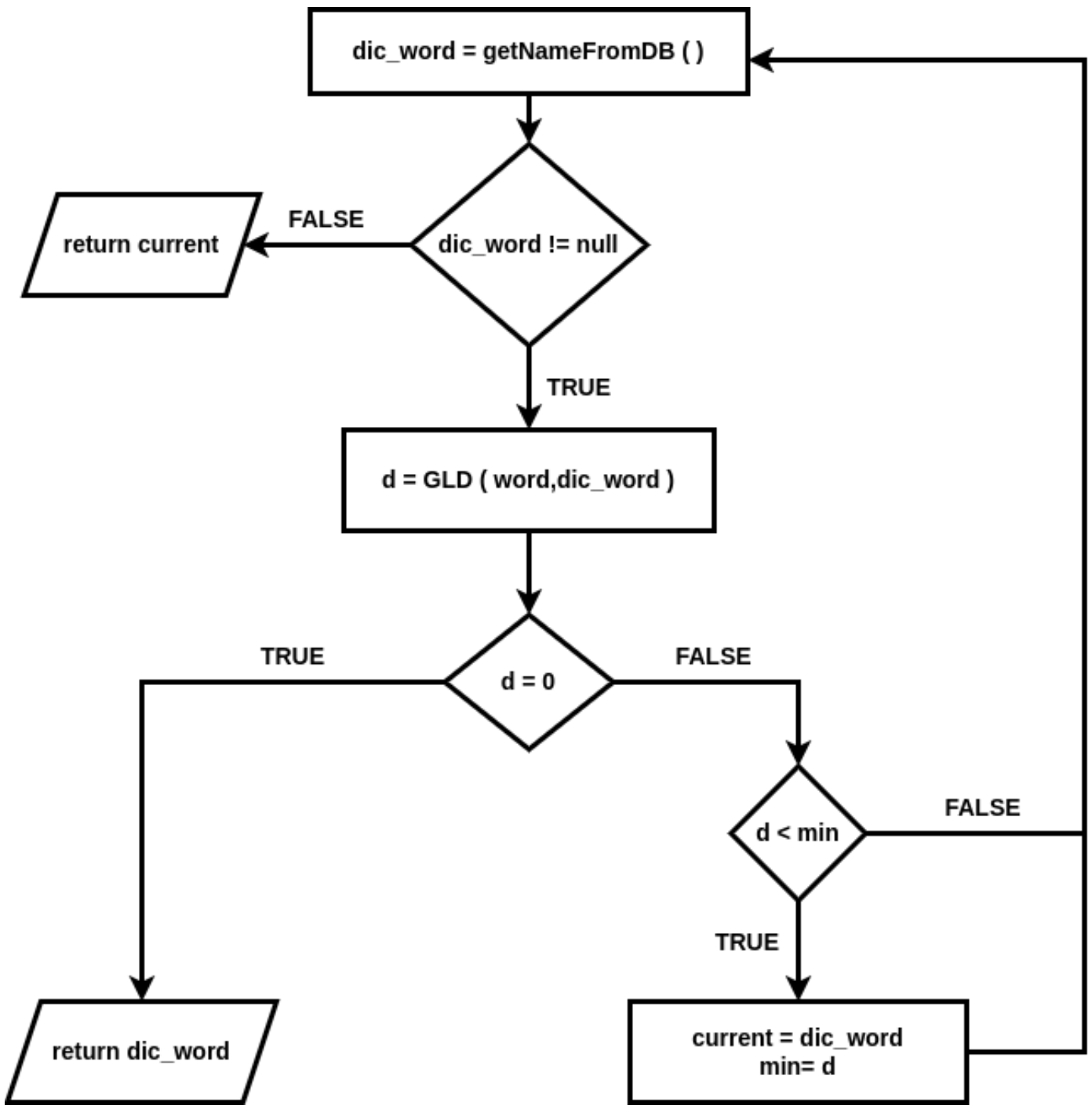

Figure 4. Matching Algorithm

In the flow diagram of Figure 4, the variables and functions are described as follows:

- Variables:
  - *word*: corresponds to the current token that the Stanford CoreNLP returns.
  - *dic_word*: is the last word read from the database of names.

- *d*: is the distance (based on the GLD) between *word* and *dic_word*.
- *current*: represents the name from the database that is the closest to *word*.
- *min*: represents the minimum distance found so far between *word* and the names in the database.

- Functions:
  - *getNameFromDB ( )*: this function reads the next name in the database.
  - *GLD (str1, str2)*: represents the GLD which determines how close or similar the two strings passed in arguments are.

In summary, this paper has proposed a new solution to the location named entities recognition problem in the context of a country where all names of regions, provinces, counties, cities, and villages are recorded in a database, and the GLD is used to avoid sensitivity to names that are misspelt. In the next section, we shall compare the solution's performance against previous ones.

## 4. RESULTS AND DISCUSSIONS

As indicated in the methodology section, we have performed an experiment to test some existing solutions on a set of 20 randomly selected texts out of the 3000 collected. The same experiment has also been done on the proposed solution and, in this section, we present and discuss the results of the experiment.

### 4.1. Results

The experiment was carried out on ChatGPT, Spacy, and Stanford CoreNLP. The results are summarised in Table 1, where the column *Expected* shows the exact list of location names contained in the text and the number of these names. Then, for each of the tools, the column *Detected* lists the locations the tool was able to recognise, and the column *Rate* gives the number of the locations recognised over the number of locations expected to be recognised.

Table 1. Experiment Results

| **#Text** | **Expected** | | **ChatGPT** | | **Spacy** | | **Stanford CoreNLP** | |
|---|---|---|---|---|---|---|---|---|
| | **Values** | **Number** | **Detected** | **Rate** | **Detected** | **Rate** | **Detected** | **Rate** |
| 1 | Komandjari<br>Gayérie | 2 | Gayérie | 1/2 | Komandjari<br>Gayérie | 2/2 | Komandjari<br>Gayérie | 2/2 |
| 2 | Oudalan<br>Zigberi<br>Markoye | 3 | Zigberi<br>Markoye | 2/3 | | 0/3 | Oudalan<br>Zigberi<br>Markoye | 3/3 |
| 3 | Seno<br>Bilakoka<br>Gorgadji | 3 | | 0/3 | Seno | 1/3 | Seno<br>Bilakoka<br>Gorgadji | 3/3 |
| 4 | Oudalan<br>Gorom | 2 | Oudalan<br>Gorom | 2/2 | Gorom | 1/2 | Oudalan<br>Gorom | 2/2 |
| 5 | Poni<br>Djigouè | 2 | La grande mosque de Djigouè | 0/2 | Poni<br>Djigouè | 2/2 | Poni<br>Djigouè | 2/2 |

| #Text | Expected | | ChatGPT | | Spacy | | Stanford CoreNLP | |
|---|---|---|---|---|---|---|---|---|
| | Values | Number | Detected | Rate | Detected | Rate | Detected | Rate |
| 6 | Oudalan<br>Deou | 2 | 2 | 1/2 | | 0/2 | Oudalan<br>Deou | 2/2 |
| 7 | Soum<br>kelbo | 2 | kelbo | 1/2 | kelbo | 1/2 | kelbo | 1/2 |
| 8 | Tuy<br>Bereba | 2 | Bereba | 1/2 | | 0/2 | | 0/2 |
| 9 | Loroum<br>Bouna<br>Titao | 3 | Bouna (12 km de Titao) | 0/3 | Loroum<br>Titao | 2/3 | Loroum<br>Bouna | 2/3 |
| 10 | Bam<br>Bourzanga | 2 | Bam<br>Bourzanga | 2/2 | Bourzanga | 1/2 | | 0/2 |
| 11 | Toboulé<br>Damba<br>Soboulé<br>Nassoumbou | 4 | Les villages de Toboulé, Damba et Soboulé (commune de Nassoumbou) | 0/4 | Toboulé<br>Damba<br>Nassoumbou | 3/4 | Toboulé<br>Damba<br>Soboulé<br>Nassoumbou | 4/4 |
| 12 | Tapoa<br>Partiaga | 2 | Partiaga | 1/2 | | 0/2 | | 0/2 |
| 13 | Bam<br>Komsilga<br>Minima<br>Zimtenga | 4 | | 0/4 | Komsilga | 1/4 | | 0/4 |
| 14 | Tapoa<br>Boungou<br>Nadiabondi | 3 | Boungou<br>Nadiabondi | 2/3 | | 0/3 | | 0/3 |
| 15 | Banwa<br>Solenzo | 2 | Solenzo | 1/2 | | 0/2 | | 0/2 |
| 16 | Tanwalbougou<br>Ougarou | 2 | Tanwalbougou<br>Ougarou | 2/2 | | 0/2 | Tanwalbougou | 1/2 |
| 17 | Kossi<br>Bourasso<br>Dedougou<br>Nouna | 4 | Bourasso<br>Axe<br>Dedougou<br>Nouna | 1/4 | | 0/4 | | 0/4 |
| 18 | Tapoa<br>Sambalgou | 2 | marché de Sambalgou | 0/2 | Tapoa | 1/2 | Tapoa | 1/2 |
| 19 | Gourma<br>Nagré | 2 | Nagré | 1/2 | Gourma | 1/2 | Gourma | 1/2 |
| 20 | Kénédougou<br>N_Dorola | 2 | | 0/2 | | 0/2 | Kénédougou<br>N_Dorola | 2/2 |

| #Text | Expected | | ChatGPT | | Spacy | | Stanford CoreNLP | |
|---|---|---|---|---|---|---|---|---|
| | Values | Number | Detected | Rate | Detected | Rate | Detected | Rate |
| | | | Average Rate | 18/50 | Average Rate | 16/50 | Average Rate | 27/50 |

As seen from Table 1, the best of the three tools is the Stanford CoreNLP with the average recognition rate of 27/50 (accuracy of 54%). However, this accuracy is low for the target type of application. The proposed solution is based on the best of the three, the **Stanford CoreNLP**, to which extensions have been made, as presented in the previous section. Table 2 shows the results of the experiment of the proposed solution over the same set of twenty texts.

Table 2. Results of the Experiment on the New Solution

| # Text | Expected | Detected | Rate |
|---|---|---|---|
| 1 | Komandjari<br>Gayérie | Komandjari<br>Gayérie | 2/2 |
| 2 | Oudalan<br>Zigberi<br>Markoye | Oudalan<br>Zigberi<br>Markoye | 3/3 |
| 3 | Seno<br>Bilakoka<br>Gorgadji | Seno<br>Bilakoka<br>Gorgadji | 2/2 |
| 4 | Oudalan<br>Gorom | Oudalan<br>Gorom | 2/2 |
| 5 | Poni<br>Djigouè | Poni<br>Djigouè | 2/2 |
| 6 | Oudalan<br>Deou | Oudalan<br>Deou | 2/2 |
| 7 | Soum<br>kelbo | Soum<br>kelbo | 2/2 |
| 8 | Tuy<br>Bereba | Tuy<br>Bereba | 2/2 |
| 9 | Loroum<br>Bouna<br>Titao | Loroum<br>Bouna<br>Titao | 3/3 |
| 10 | Bam<br>Bourzanga | Bam<br>Bourzanga | 2/2 |
| 11 | Toboulé<br>Damba<br>Soboulé<br>Nassoumbou | Toboulé<br>Damba<br>Soboulé<br>Nassoumbou | 4/4 |
| 12 | Tapoa<br>Partiaga | Tapoa<br>Partiaga | 2/2 |
| 13 | Bam<br>Komsilga<br>Minima<br>Zimtenga | Bam<br>Komsilga<br>Minima<br>Zimtenga | 4/4 |
| 14 | Tapoa<br>Boungou<br>Nadiabondi | Tapoa<br>Boungou<br>Nadiabondi | 3/3 |

| # Text | Expected | Detected | Rate |
|---|---|---|---|
| 15 | Banwa<br>Solenzo | Banwa<br>Solenzo | 2/2 |
| 16 | Tanwalbougou<br>Ougarou | Tanwalbougou<br>Ougarou | 2/2 |
| 17 | Kossi<br>Bourasso<br>Dedougou<br>Nouna | Kossi<br>Bourasso<br>Dedougou<br>Nouna | 4/4 |
| 18 | Tapoa<br>Sambalgou | Tapoa<br>Sambalgou | 2/2 |
| 19 | Gourma<br>Nagré | Gourma<br>Nagré | 2/2 |
| 20 | Kénédougou<br>N_Dorola | Kénédougou | 1/2 |
| | **Average Rate** | | **49/50** |

The new solution has a recognition rate of 49/50 (accuracy of 98%).

## 4.2. Discussions

The accuracy of ChatGPT, Spacy, Stanford CoreNLP, and the proposed solution on the test set is given in Figure 5. In terms of accuracy, the proposed solution is outstanding, giving more confidence to users who want to extract information from social network texts.

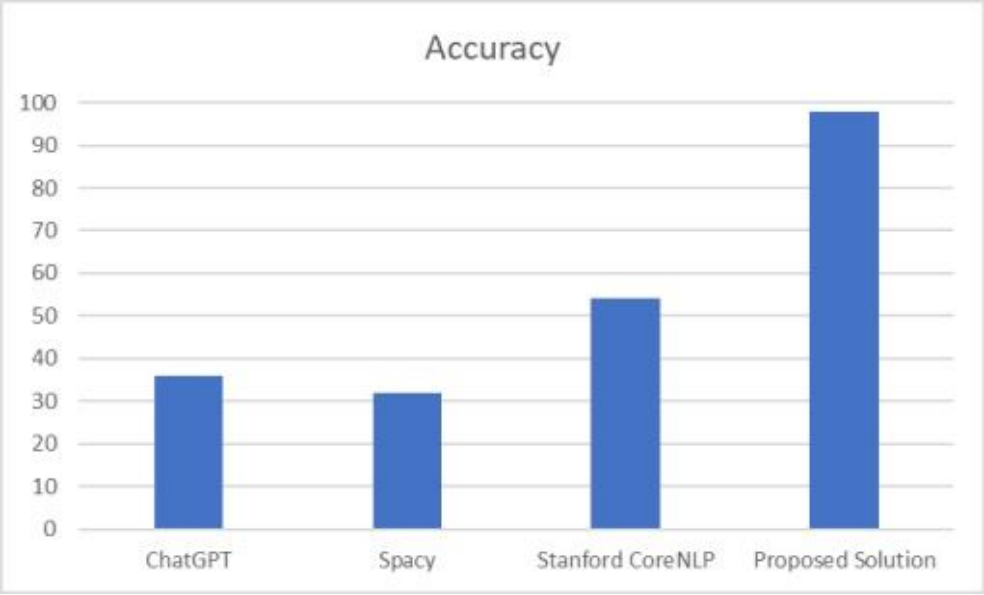

Figure 5. Accuracy of the Solutions

The low performance of NER tools compared to our proposed solution can be explained by two reasons. Firstly, the French language is known to have less publicly available labelled data that participate in the training of the NER tools [25]. Secondly, the social network texts are often poorly formulated, with grammatical and syntax errors which make the context difficult to understand by the tools and thus failing to recognise location entities. Besides the accuracy, we have also compared the speed (execution time) of the new solution to the others. The execution time on a Core i5 CPU @2.3 GHZ, 12 GB RAM computer was respectively 55 seconds, 3 seconds, 59 seconds, and 293 seconds. Despite this higher execution time of the proposed solution, we still recommend it because the gain in accuracy outweighs the speed deficit.

## 5. CONCLUSION

The objective of this first phase of the research project, which was to build a high-accuracy location name recognition system, has been achieved. The proposed solution has an accuracy of 98%, which no other tool, according to our knowledge, has been able to reach. The new solution is both an extension and a simplification of the Stanford CoreNLP: a simplification because we have reduced the pipeline to the tokenisation phase and an extension because we have introduced the pre-processing and location recognizer steps. It is also important to note that the gain in accuracy did result in significant overhead in execution time. However, for most applications, execution time is not a crucial factor.

While the result of this first phase is outstanding, the proposed solution is of little interest if the project's subsequent phases are not accomplished. Other information to extract from the internet and social network texts include dates and terrorist actions. In the project's next phase, we will address the extraction of these types of information to provide a complete solution.

### ACKNOWLEDGEMENTS

This research has not received any specific support to be acknowledged.

## AUTHORS

**Lossan Bonde** is a PhD holder since 2006 from the University of Science and Technologies of Lille, France. He is currently assistant professor of computer science at the Adventist University of Africa, Nairobi, Kenya. His research interests are in Artificial Intelligence and the Internet of Things with special focus on building NLP solutions for real life problems.

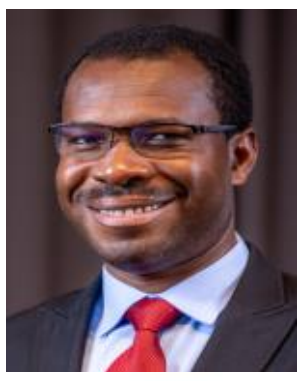

**Severin Dembele** has completed a master's degree in computer Science with specialisation in Decision Support Systems at the Public University, Nazi Boni, Bobo-Dioulasso, Burkina Faso. He is in the process of starting his PhD studies in the field of NLP which is also his current research interest.

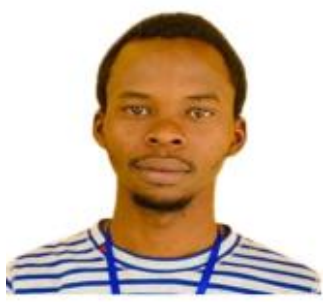